\documentclass{article}
\pdfoutput=1
\usepackage[preprint]{neurips_2023}
\usepackage[utf8]{inputenc} %
\usepackage{booktabs}
\usepackage{booktabs}
\usepackage{diagbox} 

\usepackage[T1]{fontenc}    %
\usepackage{xurl}
\usepackage{hyperref}            %
\usepackage{breakurl}          %
\usepackage{amsfonts}       %
\usepackage{amsmath}       %
\usepackage{nicefrac}       %
\usepackage{microtype}      %
\usepackage{lipsum}
\usepackage{graphicx}
\usepackage{makecell}
\usepackage{wrapfig}
\usepackage[size=small]{caption}
\usepackage{mathtools}
\usepackage{amssymb}
\usepackage{amsthm}
\usepackage[algoruled,boxed,lined,noend]{algorithm2e}
\usepackage{adjustbox} 
\usepackage{float}
\usepackage{pifont}
\usepackage{subcaption}
\usepackage{fontawesome}
\usepackage{multirow}
\usepackage{rotating}
\usepackage{array}    
\usepackage{lineno}
\usepackage{longtable}
\usepackage[table,dvipsnames]{xcolor}
\usepackage[numbers]{natbib}
\usepackage{textcomp}
\usepackage{ulem}
\usepackage{color}
\usepackage[most]{tcolorbox}
\usepackage{xcolor}

\definecolor{qual-fig-green}{RGB}{0,144,11}
\definecolor{qual-fig-red}{RGB}{238,0,0}
\definecolor{qual-fig-purple}{RGB}{153,51,255}

\definecolor{answergreen}{RGB}{222,255,222}
\definecolor{promptblue}{RGB}{235,245,255}   
\definecolor{promptgreen}{RGB}{235,255,235}  
\definecolor{promptgray}{RGB}{248,248,248}

\title{Is ChatGPT-5 Ready for Mammogram VQA?}

\author{
Qiang Li$^{1}$ \quad Shansong Wang$^{1}$   \quad Mingzhe Hu$^{1}$  \quad Mojtaba Safari$^{1}$ \quad Zachary Eidex$^{1}$ \quad \textbf{Xiaofeng Yang}$\textsuperscript{1 \faEnvelope}$ 
\vspace{3mm}
\\
$^1$Department of Radiation Oncology, Winship Cancer Institute, Emory University School of Medicine\\
\faEnvelope \quad Corresponding author: xiaofeng.yang@emory.edu \\
}

\begin{document}
\maketitle

\begin{abstract}
Mammogram visual question answering (VQA) integrates image interpretation with clinical reasoning and has potential to support breast cancer screening. We systematically evaluated the GPT-5 family and GPT-4o model on four public mammography datasets (EMBED, InBreast, CMMD, CBIS-DDSM) for BI-RADS assessment, abnormality detection, and malignancy classification tasks. GPT-5 consistently was the best performing model but lagged behind both human experts and domain-specific fine-tuned models. On EMBED, GPT-5 achieved the highest scores among GPT variants in density (56.8\%), distortion (52.5\%), mass (64.5\%), calcification (63.5\%), and malignancy (52.8\%) classification. On InBreast, it attained 36.9\% BI-RADS accuracy, 45.9\% abnormality detection, and 35.0\% malignancy classification. On CMMD, GPT-5 reached 32.3\% abnormality detection and 55.0\% malignancy accuracy. On CBIS-DDSM, it achieved 69.3\% BI-RADS accuracy, 66.0\% abnormality detection, and 58.2\% malignancy accuracy. Compared with human expert estimations, GPT-5 exhibited lower sensitivity (63.5\%) and specificity (52.3\%). While GPT-5 exhibits promising capabilities for screening tasks, its performance remains insufficient for high-stakes clinical imaging applications without targeted domain adaptation and optimization. However, the tremendous improvements in performance from GPT-4o to GPT-5 show a promising trend in the potential for general large language models (LLMs) to assist with mammography VQA tasks.

\end{abstract}

\section{Introduction}
Breast cancer is the most frequently diagnosed malignancy among women worldwide and a leading cause of cancer-related death \cite{kim2025global}. Mammography remains the cornerstone of large-scale breast cancer screening, providing a critical tool for identifying abnormal tissue changes. Interpreting these images, however, is a complex and time-consuming task that demands specialized expertise and extensive clinical experience. The variability in breast tissue patterns, combined with the often subtle presentation of early-stage cancers, makes accurate reading challenging \cite{lehman2017national}. Even among expert radiologists, performance can vary significantly \cite{Gilbert2006}. These inherent difficulties have spurred the advancement of artificial intelligence technologies aimed at supporting radiologists, improving diagnostic accuracy, and promoting greater consistency in interpretations. \cite{tu2024towards,chen2023performance,wu2019deep,chougrad2018deep}.

Given the complexity of medical image interpretation, zero-shot application of large language or vision language models without any domain-specific fine-tuned may yield suboptimal results\cite{he2025parameter,safari2025performancegpt5braintumor}. AI models fine-tuned on medical imaging data have demonstrated strong performance across multiple radiology modalities and downstream tasks, including classification, detection, and segmentation, often surpassing traditional CAD systems in sensitivity and specificity~\cite{tajbakhsh2016convolutional}.
 Large vision-language models further extend these advances, and when fine-tuned on domain-specific medical images, have demonstrated strong performance across several imaging modalities, including radiography, pathology, and retinal imaging. Medical visual question answering (VQA) builds on this foundation by combining image understanding with natural language reasoning, enabling models to answer clinically framed questions directly~\cite{wang2025capabilities,he2020pathvqa,zhang2023pmc}. This paradigm offers a practical framework to test whether general-purpose multimodal models like ChatGPT-5 can interpret mammograms in ways that align with clinical decision-making.

Given recent reports of GPT-5's impressive performance gains over previous generations, we ask "Is ChatGPT-5 ready for mammogram VQA?" We evaluate the zero-shot performance of ChatGPT-5 with multimodal capabilities on mammogram interpretation tasks. Using a diverse set of U.S.\ screening mammograms, we assess its ability to classify malignancy and related findings from single-view images. We construct VQA items from four public datasets: \textbf{EMBED}~\cite{jeong2023emory}, \textbf{INBreast}~\cite{moreira2012inbreast}, \textbf{CMMD}~\cite{cui2021chinese}, and \textbf{CBIS-DDSM}~\cite{sawyer2016curated},then tailoring the metadata and annotations into standardized templates. To control external factors, we use a fixed zero-shot prompt and evaluate each case on a single mammogram view without multi-view or clinical context.

\section{Methodology}
\subsection{Datasets}
To evaluate GPT\mbox{-}5 on digital X\mbox{-}ray mammography reasoning, we derive clinically relevant VQA items from the structured (panel) labels available in four public mammography datasets: EMBED~\cite{jeong2023emory}, INBreast~\cite{moreira2012inbreast}, CMMD~\cite{cui2021chinese}, and CBIS\mbox{-}DDSM~\cite{sawyer2016curated}. We harmonize dataset\mbox{-}specific metadata and clinical annotations (laterality, view, image type, BI\mbox{-}RADS breast density and assessment, lesion types/attributes, and biopsy\mbox{-}confirmed pathology) into a common schema and convert them into standardized question templates. These include lesion presence/type (mass, calcification, none, both), malignancy classification (benign vs.\ malignant), breast density (A--D), and ROI\mbox{-}conditioned attribute questions (e.g., mass shape/margins or calcification distribution).
\begin{itemize}
\item \textbf{EMBED}~\cite{jeong2023emory} contains 3.4 million screening and diagnostic images from 110{,}000 patients collected between 2013 and 2020 and focuses on generalizabity across ethnic groups by providing a balanced representation of Black and White women~\cite{jeong2023emory}. It includes 2D, synthetic 2D (C-view), and 3D digital breast tomosynthesis (DBT) images, along with 60{,}000 annotated lesions linked to structured imaging descriptors and pathology-confirmed outcomes across six severity classes. In this study, we use the publicly available subset, which represents 20\% of the 2D portion of the dataset.
\item \textbf{InBreast}~\cite{moreira2012inbreast} comprises 115 cases (410 images), including 90 cases from women with bilateral breast involvement (four images per case) and 25 cases from mastectomy patients (two images per case). It encompasses a variety of lesion types, such as masses, calcifications, asymmetries, and distortions.
\item \textbf{CMMD}~\cite{cui2021chinese} is a publicly available mammography dataset curated by the Cancer Hospital of the Chinese Academy of Medical Sciences~\cite{cui2021chinese}. It contains full-field digital mammography (FFDM) images from 1{,}775 patients collected between 2012 and 2016, accompanied by clinical information such as age, pathology results, and lesion type. Images are labeled at the breast level as benign or malignant based on pathology-confirmed outcomes. CMMD provides a valuable resource for developing and evaluating computer-aided diagnosis methods on a non-Western population.
\item \textbf{CBIS-DDSM}~\cite{sawyer2016curated} is a database of 2,620 scanned film mammography studies. It contains normal, benign, and malignant cases with verified pathology information. Using CBIS-DDSM, we construct VQA items focused on lesion-type recognition (mass vs. calcification vs. none), ROI-aware localization/attributes, and pathology-linked questions where available. To ensure consistent evaluation coverage across datasets and question categories, we adopt a balanced sampling strategy: for each question type and answer category, we randomly sample an equal number of items, thereby avoiding bias toward over-represented classes in large datasets while keeping the total evaluation set size tractable.
\end{itemize}
\subsection{VQA generation}
The visual question answering (VQA) pairs were constructed from four publicly available mammography datasets: EMBED, InBreast, CMMD, and CBIS-DDSM. For each case, questions were automatically generated from the structured panel data and metadata accompanying the images, including patient information, imaging annotations, lesion type, BI-RADS density category, and biopsy-confirmed pathology. This design yields questions that target well-defined clinical tasks, such as “What is the BI-RADS breast density?” or “Is the imaging finding suggestive of malignancy?”, ensuring a one-to-one correspondence between the query, its answer, and the gold-standard label. By avoiding free-text questions based on subjective interpretation of image texture, morphology, or density patterns, the approach reduces variability and enhances reproducibility in automated evaluation. However, the absence of texture-oriented, descriptive queries limits the direct assessment of visual reasoning processes, positioning the evaluation as a test of structured clinical label inference rather than nuanced image-based reasoning.

\subsection{Prompt Design}
We evaluate GPT-5 using a zero-shot chain-of-thought (CoT) prompting strategy~\cite{openai2025gpt5}, implemented as a concise two-turn dialogue. In the first turn, a system message establishes the medical domain context, and the user presents the question, explicitly triggering step-by-step reasoning with the cue - \textbf{Let’s think step by step.} For multimodal questions, all associated images are provided in this message as \texttt{image\_url} entries, enabling the model to jointly process visual and textual information within a single reasoning phase. The model responds with an unconstrained explanatory rationale (\textit{prediction\_rationale}) without selecting a final answer.

\begin{figure}[htbp]
\centering
\begin{tcolorbox}[
    colback=promptgray,
    colframe=black!50!black,
    colbacktitle=white!50!black,
    title=A Sample from EMBED dataset,
    fontupper=\normalsize,
    enhanced,
    overlay={
        \node[anchor=north east, inner sep=3pt, yshift=-15pt] 
        at (frame.north east) {\includegraphics[width=0.25\textwidth]{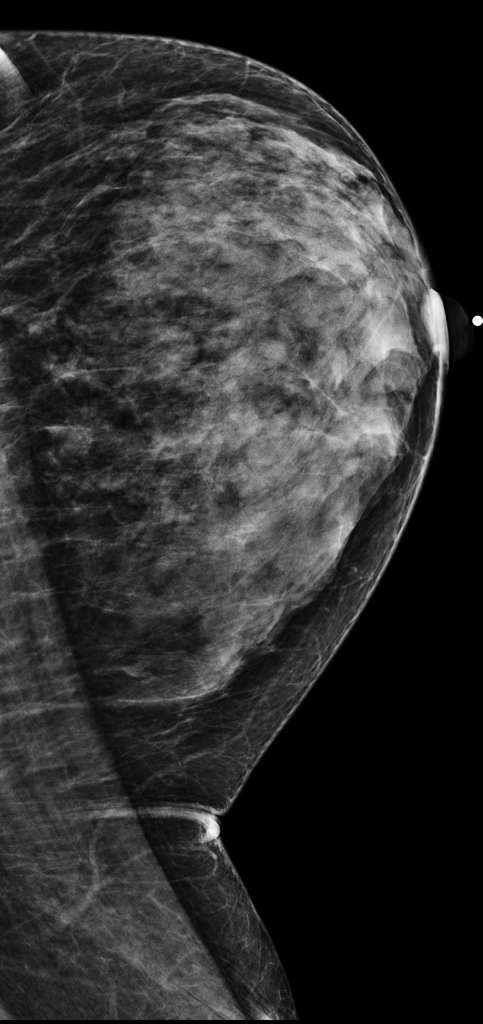}};
    } 
]
{\linespread{0.8}\selectfont
\begin{verbatim}
[
  {
    "role": "system",
    "content": "You are a helpful medical assistant."
  },
  {
    "role": "user",
    "content": [
      {
        "type": "text",
        "text": "
Q: What is the BI-RADS breast density of this image?          
Answer Choices: 
(A)BI-RADS A (B)BI-RADS B (C)BI-RADS C (D)BI-RADS D
            A: Let's think step by step."
      },
      {
        "type": "image_url",
        "image_url": { "url": "images/MM-4307.png" }
      }
    ]
  },
  {
    "role": "assistant",
    "content": "{ASSISTANT_RATIONALE}"
  },
  {
    "role": "user",
    "content": "Therefore, among {start} through {end}, the answer is"
  },
  {
    "role": "assistant",
    "content": "D"
  }
]
\end{verbatim}
}
\end{tcolorbox}
\caption{An example multimodal prompt from the EMBED dataset. The model is tasked with provided the correct answer given a mammogram and multiple choice question.}
\label{fig:promptVQAsample}
\end{figure}

In the second turn, the user issues a convergence instruction:\textbf{Therefore, among A through \{\texttt{END\_LETTER}\}, the answer is} - where \{\texttt{END\_LETTER}\} denotes the last available option label. At this stage, the model outputs only a single choice letter (\textit{prediction}). This separation ensures that the reasoning process and final decision are elicited in distinct conversational steps. The jsonl templates for both text-only and image-augmented formats follow this protocol, using placeholders \{\texttt{QUESTION\_TEXT}\}, \{\texttt{END\_LETTER}\}, \{\texttt{IMAGE\_URL\_1}\}, \{\texttt{ASSISTANT\_RATIONALE}\}, and \{\texttt{ASSISTANT\_FINAL}\}. The prompting design template for the mammogram VQA task is illustrated in Fig.~\ref{fig:promptVQAsample}.

\label{fig:promptVQA}

\section{Results}

\subsection{Performance of GPTs on public mammograms dataset screening}
Table \ref{tab:embed_results1}-\ref{tab:cbis_ddsm_results4} summarizes the screening performance (\%) of GPT family models on multiple public mammography datasets, including EMBED, InBreast, CMMD, and CBIS-DDSM. Across datasets, GPT-5 consistently outperforms its smaller variants (GPT-5-mini, GPT-5-nano) and the previous-generation GPT-4o (GPT-4o-2024-11-20 version), particularly in tasks requiring fine-grained lesion characterization such as mass and calcification detection. While GPT-4o demonstrates reasonable accuracy in several settings, the performance gap relative to GPT-5 is more pronounced in multi-class tasks (e.g., BI-RADS classification) and malignancy assessment. 
Across four public mammography datasets (EMBED, InBreast, CMMD, CBIS-DDSM), GPT-5 shows measurable but still limited screening performance compared to specialized medical AI systems and human experts. On the EMBED dataset, GPT-5 achieves the highest performance among GPT variants (e.g., Density: 56.8\%, Distortion: 52.5\%, Mass: 64.5\%, Calcification: 63.5\%, Malignancy: 52.8\%), outperforming GPT-5-mini, GPT-5-nano, and GPT-4o, but still lagging behind pretrained fine-tuned SOTA models.

On the InBreast dataset, GPT-5 attains 36.9\% BI-RADS accuracy, 45.9\% for abnormality detection, and 35.0\% for malignancy classification, far below the 90.6\% malignancy accuracy of MRSN. In CMMD, GPT-5 records 32.3\% abnormality detection and 55.0\% malignancy accuracy, versus 79.7\% from HybMNet.

On CBIS-DDSM, GPT-5 achieves 69.3\% BI-RADS accuracy, 66.0\% abnormality detection, and 58.2\% malignancy accuracy. Compared to human expert estimations (sensitivity 86.9\%, specificity 88.9\%), GPT-5's sensitivity drops to 63.5\% ($\downarrow$23.4\%) and specificity to 52.3\% ($\downarrow$36.6\%). Specialized models such as PHYSnet (82.0\% malignancy accuracy) and ResNet18-S896 (79.6\%) substantially outperform GPT-5.

Overall, GPT-5 consistently surpasses smaller GPT variants but remains notably behind both human experts and domain-optimized AI models, indicating substantial room for improvement in domain adaptation, reasoning transparency, and uncertainty calibration for high-stakes clinical imaging tasks.

These results highlight the potential of large-scale, instruction-tuned multimodal language models for zero-shot mammography interpretation, though specialized fine-tuned state-of-the-art (SOTA) models still achieve superior results on certain tasks, indicating room for further domain adaptation.

{\small
\begin{table}[ht!]
\centering
\caption{Performance comparison between GPTs and pretrained fine-tuned SOTA models on the EMBED dataset.}
\label{tab:embed_results1}
\begin{tabular}{lccccc}
\toprule
\diagbox[dir=SE,width=10em]{\textbf{Model}}{\textbf{Tasks}} 
  & \textbf{Density} & \textbf{Distortion} & \textbf{Mass} & \textbf{Calcification} & \textbf{Malignancy} \\
\midrule
GPT-5              & 56.8 & 52.5 & 64.5 & 63.5 & 52.8 \\
GPT-5-mini         & 34.3 & 53.5 & 60.8 & 57.3 & 47.3 \\
GPT-5-nano         & 24.8 & 53.5 & 52.3 & 51.5 & 47.8 \\
GPT-4o             & 24.3 & 20.0 & 50.0 & 44.3 & 42.5 \\
\hline
Mammo-CLIP (ViT-B/16) \cite{chen2024mammo} & - & - & - &  - & 79.0  \\
Mammo-CLIP (ViT-L/14) \cite{chen2024mammo} &- & - &  - & - & 82.3  \\
\bottomrule
\end{tabular}
\end{table}
}

{\small
\begin{table}[ht!]
\centering
\caption{Performance comparison between GPTs and pretrained fine-tuned SOTA models on the InBreast dataset.}
\label{tab:inbreast_results2}
\begin{tabular}{lccc}
\toprule
\diagbox[dir=SE,width=10em]{\textbf{Model}}{\textbf{Tasks}} 
  & \textbf{BI-RADS} & \makecell{\textbf{Abnormality} \\ calcification and mass} & \textbf{Malignancy} \\
\midrule
GPT-5              & 36.9 & 45.9  & 35.0 \\
GPT-5-mini         & 28.1 & 49.2  & 40.0 \\
GPT-5-nano         & 17.6 & 37.8  & 21.5 \\
GPT-4o             & 23.7 & 36.1  & 30.4 \\
\hline
MRSN \cite{sun2025multi} & - & - & 90.6 \\
GGP \cite{sun2025multi} & - & - & 88.5\\
\bottomrule
\end{tabular}
\end{table}
}
{\small
\begin{table}[ht!]
\centering
\caption{Performance comparison between GPTs and pretrained fine-tuned SOTA models on the CMMD dataset.}
\label{tab:cmmd_results3}
\begin{tabular}{lccc}
\toprule
\diagbox[dir=SE,width=10em]{\textbf{Model}}{\textbf{Tasks}} 
  & \textbf{BI-RADS} 
  & \makecell{\textbf{Abnormality} \\ calcification and mass} 
  & \textbf{Malignancy} \\
\midrule
GPT-5              & - & 32.3 & 55.0 \\
GPT-5-mini         & - & 42.3 & 63.3 \\
GPT-5-nano         & - & 34.0 & 52.7 \\
GPT-4o             & - & 38.5 & 48.5 \\
\hline
HybMNet \cite{chen2025enhancing}            & - & - & 79.7 \\
GMIC \cite{chen2025enhancing}               & - & - & 73.6 \\
\bottomrule
\end{tabular}
\end{table}
}

{\small
\begin{table}[ht!]
\centering
\caption{Performance comparison between GPTs and pretrained fine-tuned SOTA models on the CBIS-DDSM dataset.}
\label{tab:cbis_ddsm_results4}
\begin{tabular}{lccc}
\toprule
\diagbox[dir=SE,width=10em]{\textbf{Model}}{\textbf{Tasks}} 
  & \textbf{BI-RADS} 
  & \makecell{\textbf{Abnormality} \\ calcification and mass} 
  & \textbf{Malignancy} \\
\midrule
GPT-5              & 69.3 & 66.0 & 58.2 \\
GPT-5-mini         & 43.6 & 53.3 & 43.5 \\
GPT-5-nano         & 20.2 & 41.8 & 39.0 \\
GPT-4o             & 28.4 & 46.2 & 40.0 \\
\hline
MRSN \cite{sun2025multi} & - & - & 77.8 \\  
Single view baseline CNN \cite{gudhe2024multi} & - & - & 71.0 \\
Single view evidential CNN \cite{gudhe2024multi} & - & - & 72.6 \\
PHYSnet \cite{lopez2022multi} & - & - & 82.0 \\
ResNet18-S896 \cite{wei2022beyond} & - & - & 79.6 \\
\bottomrule
\end{tabular}
\end{table}
}

\subsection{Comparison with human experts}

Table~\ref{tab:medxpertqa-text-mm} compares the malignant screening performance of different GPT model variants with human expert estimations on the CBIS-DDSM dataset. Human readers achieve a sensitivity of 86.9\% and specificity of 88.9\%, whereas all GPT models fall substantially short in both metrics. GPT-5 attains the highest accuracy among model variants (58.2\%) but still shows marked reductions in sensitivity ($-23.4\%$) and specificity ($-36.6\%$) relative to human performance. Lighter GPT-5 variants (mini and nano) and GPT-4o exhibit even larger performance gaps, with sensitivity drops exceeding 50\% and specificity deficits up to 26.0\%. These findings reflect the intrinsic challenges of mammogram interpretation, where diagnostic decisions hinge on subtle, low-contrast imaging features, and emphasize that without domain-specific fine-tuning, current general-purpose multimodal LLMs remain far from achieving expert-level screening accuracy.

\begin{table}[ht!]
\centering
\caption{Comparison with Human Experts on CBIS-DDSM (Malignant Screening).}
\label{tab:medxpertqa-text-mm}
\begin{tabular}{lccc}
\hline
 & Sensitivity & Specificity & ACC \\
\hline
Human Estimation \cite{lehman2017national,wei2022beyond} 
& 86.9 & 88.9 & -- \\
\hline
GPT-5
& \makecell{63.5\\ (\textcolor{red}{$\downarrow$23.4\%})}
& \makecell{52.3\\ (\textcolor{red}{$\downarrow$36.6\%})}
& 58.2 \\
GPT-5-mini
& \makecell{35.8\\ (\textcolor{red}{$\downarrow$51.1\%})}
& \makecell{62.9\\ (\textcolor{red}{$\downarrow$26.0\%})}
& 44.9 \\
GPT-5-nano
& \makecell{34.9\\ (\textcolor{red}{$\downarrow$52.0\%})}
& \makecell{67.5\\ (\textcolor{red}{$\downarrow$21.4\%})}
& 51.2 \\
GPT-4o
& \makecell{33.3\\ (\textcolor{red}{$\downarrow$53.6\%})}
& \makecell{66.7\\ (\textcolor{red}{$\downarrow$22.2\%})}
& 40.0 \\
\hline
\end{tabular}
\end{table}

\subsection{Case Studies on GPT Errors in Mammogram VQA}
We selected four representative examples, two correctly classified and two misclassified, to qualitatively examine GPT-5’s decision-making in BI-RADS density assessment and malignancy detection tasks. These cases illustrate the model’s strengths in recognizing prototypical imaging features and its weaknesses in borderline density categorization and atypical lesion interpretation, shown in \begin{table}[htbp]
\centering
\caption{Representative GPT-5 predictions: two correct and two incorrect cases in BI-RADS density and malignancy classification.}
\begin{tabular}{p{0.23\linewidth} p{0.23\linewidth} p{0.23\linewidth} p{0.23\linewidth}}
\toprule
\textbf{Correct: Malignancy (Yes)} & \textbf{Correct: Malignancy (Yes)} & \textbf{Error: Density D$\rightarrow$C} & \textbf{Error: Malignancy B$\rightarrow$A} \\
\midrule
\includegraphics[width=\linewidth]{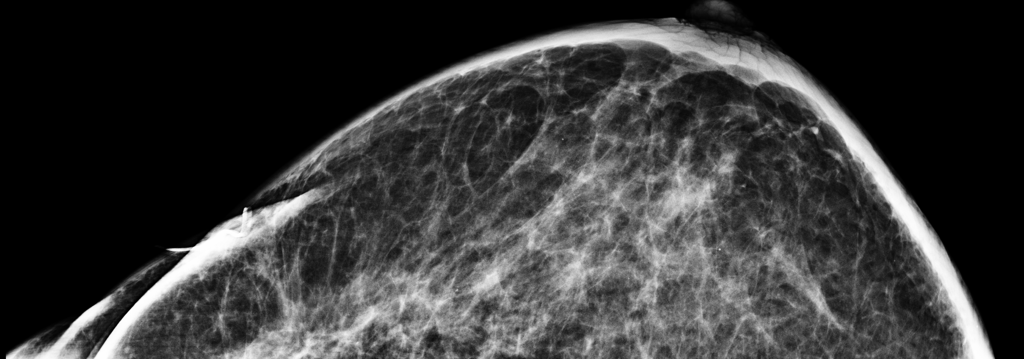} &
\includegraphics[width=\linewidth]{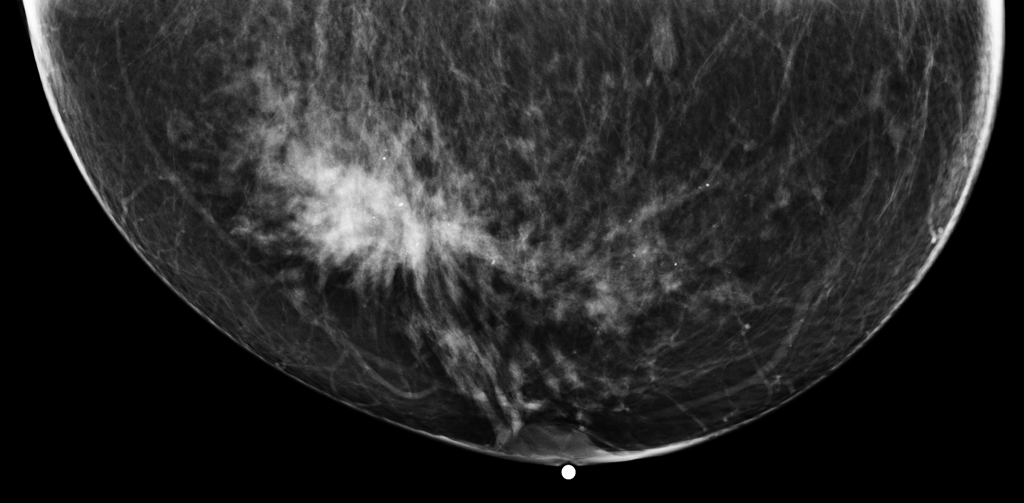} &
\includegraphics[width=\linewidth]{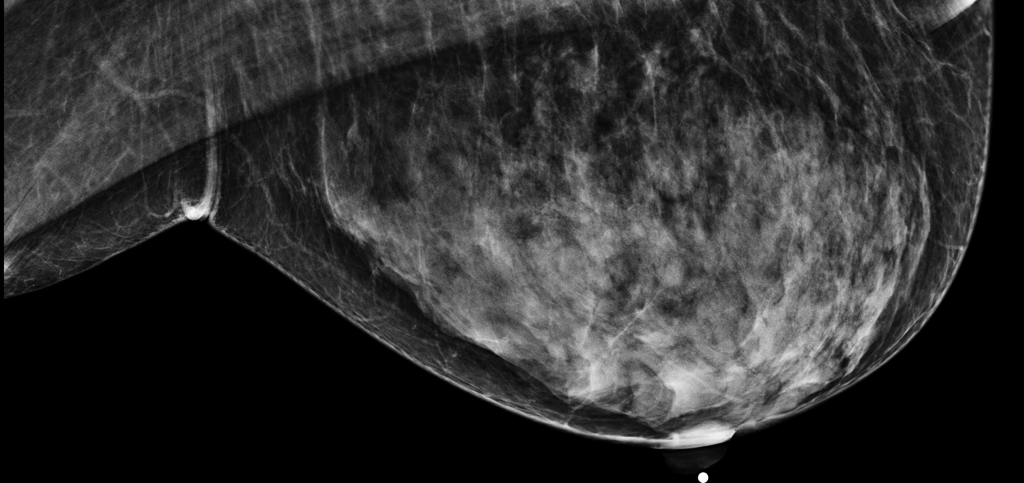} &
\includegraphics[width=\linewidth]{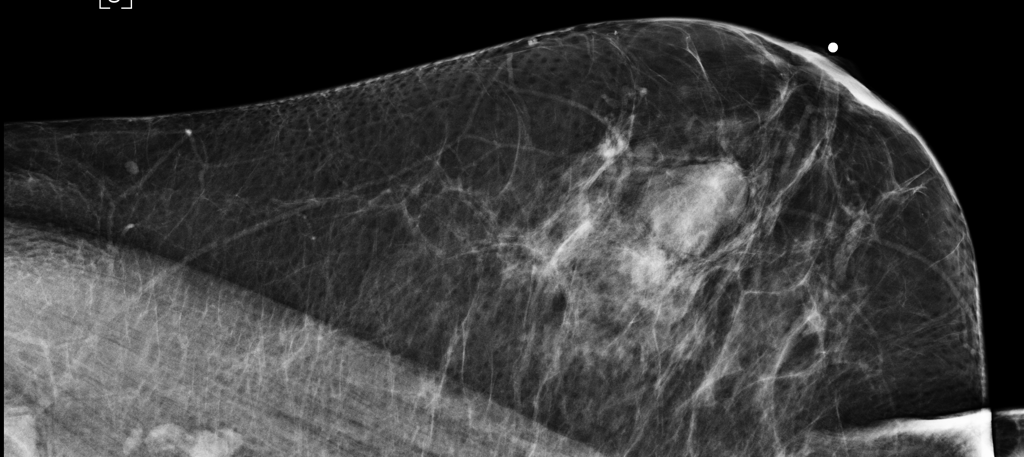} \\
\small \textbf{GT:} Yes \newline \textbf{Pred:} Yes \newline \textbf{Rationale:} The mammogram does not show a discrete spiculated mass, suspicious pleomorphic/linear branching calcifications, or convincing architectural distortion. Findings appear consistent with normal fibroglandular tissue without specific features of malignancy.&
\small \textbf{GT:} Yes \newline \textbf{Pred:} Yes \newline \textbf{Rationale:} A) Yes. The image shows an irregular spiculated mass with architectural distortion and suspicious calcifications—features suggestive of malignancy. &
\small \textbf{GT:} D \newline \textbf{Pred:} C \newline \textbf{Rationale:} The breast shows large regions of fibroglandular tissue with some interspersed fat—heterogeneously dense, not almost entirely fatty (A/B) nor uniformly/extremely dense (D). &
\small \textbf{GT:} No \newline \textbf{Pred:} Yes \newline \textbf{Rationale:} (A) Yes. The mass is irregular and spiculated with associated architectural distortion—features that are highly suspicious for malignancy. \\
\bottomrule
\end{tabular}
\label{tab:correct_incorrect_examples}
\end{table}

Correct cases (MM-11387, MM-9467):
In BI-RADS density classification and benign/malignant discrimination, GPT-5 successfully identified classic radiological features.
In the correct density example, the model accurately recognized an almost entirely dense breast.
In the correct malignancy example, the model correctly identified a lesion with an irregular spiculated mass, architectural distortion, and suspicious calcifications (hallmark features of malignancy).

Error cases (MM-4307, MM-20757):
For density assessment errors, the model consistently misclassified BI-RADS category D (extremely dense) as category C (heterogeneously dense), indicating a tendency to underestimate overall density in highly dense breasts.
For malignancy assessment errors, benign structural changes were misclassified as malignant (overdiagnosis), suggesting the model is prone to false-positive interpretations when confronted with architectural distortion or irregular mass margins lacking other corroborating malignant features.

\section{Discussion}

Our evaluation indicates that ChatGPT-5, as a general-purpose multimodal LLM, struggles with complex, domain-specific queries in mammography. This limitation is likely due to the model’s lack of fine-tuning on high-resolution grayscale breast imaging and the absence of explicit adaptation to structured clinical tasks such as BI-RADS classification or lesion characterization. Mammogram interpretation itself is inherently challenging, even for experienced radiologists. Diagnostic decisions often hinge on subtle, low-contrast features, such as microcalcification clusters, spiculated mass margins, or architectural distortions that can be obscured by overlapping fibroglandular tissue, particularly in dense breasts, and further complicated by variations in acquisition parameters and image quality. Accurate assessment requires not only fine-grained visual discrimination but also the integration of patient history, comparison with prior examinations, and adherence to standardized reporting frameworks like BI-RADS. These factors underscore why mammographic VQA is especially difficult for general-purpose AI systems without targeted domain adaptation and large-scale, modality-specific training.

\subsection{Limitations}
This study has several limitations. First, both the evaluation data and the comparison model outputs were drawn from the same datasets, but a fully unified test set across all models was not feasible. For datasets with extremely large sample sizes, it was impractical to perform exhaustive evaluation on the entire test split. Instead, we adopted a balanced sampling strategy, evenly distributing samples across all question types and answer categories to ensure representative coverage. As such, the reported results should be regarded as indicative rather than exhaustive. Second, all experiments were conducted in a zero-shot setting using ChatGPT-5 without domain-specific fine-tuning or prompt optimization for mammography. This design allows us to assess the model’s intrinsic, out-of-the-box capabilities, but may underestimate the achievable performance with targeted adaptation. Third, the evaluation was limited to a fixed set of question templates derived from structured labels, which may not capture the full variability and nuance of real-world clinical questioning. Fourth, while several SOTA methods have reported strong results on the same datasets, direct comparison remains challenging due to differences in experimental setups, data preprocessing pipelines, and evaluation protocols. Finally, our analysis focused primarily on accuracy-based metrics; other important aspects such as reasoning transparency, uncertainty calibration, and systematic error characterization were not explored.

\section{Conclusion}
In this study, we preliminarily evaluated the zero-shot performance of the ChatGPT-5 family and GPT-4o model on clinically relevant VQA tasks in digital and digitized mammography, using four diverse public datasets (EMBED, InBreast, CMMD, and CBIS-DDSM). ChatGPT-5 lags behind physicians and specialized, fine-tuned SOTA models on most tasks and is prone to false-positive interpretations when confronted with architectural distortion or irregular mass margins. While not ready for clinical use, GPT-5 achieved considerable improvement over the GPT-4o model, so we are optimistic about the potential for generalist multimodal LLMs to improve patient care in mammography VQA tasks.

\section*{Code Availability}
The code used in this study is publicly available at 
\href{https://github.com/qiangCS/GPT-5-Evaluation}{\texttt{GPT-5-Evaluation}}.

\bibliographystyle{unsrt}
\bibliography{newbib}

\begin{thebibliography}{10}

\bibitem{kim2025global}
Joanne Kim, Andrew Harper, Valerie McCormack, Hyuna Sung, Nehmat Houssami,
  Eileen Morgan, Miriam Mutebi, Gail Garvey, Isabelle Soerjomataram, and
  Miranda~M Fidler-Benaoudia.
\newblock Global patterns and trends in breast cancer incidence and mortality
  across 185 countries.
\newblock {\em Nature Medicine}, pages 1--9, 2025.

\bibitem{lehman2017national}
Constance~D Lehman, Robert~F Arao, Brian~L Sprague, Janie~M Lee, Diana~SM
  Buist, Karla Kerlikowske, Louise~M Henderson, Tracy Onega, Anna~NA Tosteson,
  Garth~H Rauscher, et~al.
\newblock National performance benchmarks for modern screening digital
  mammography: update from the breast cancer surveillance consortium.
\newblock {\em Radiology}, 283(1):49--58, 2017.

\bibitem{Gilbert2006}
Fiona~J. Gilbert, Susan~M. Astley, Michael~A. McGee, Mary~G. Gillan,
  Christopher~R. Boggis, Paul~M. Griffiths, and Stephen~W. Duffy.
\newblock Single reading with computer-aided detection and double reading of
  screening mammograms in the united kingdom national breast screening program.
\newblock {\em Radiology}, 241(1):47--53, October 2006.

\bibitem{tu2024towards}
Tao Tu, Shekoofeh Azizi, Danny Driess, Mike Schaekermann, Mohamed Amin,
  Pi-Chuan Chang, Andrew Carroll, Charles Lau, Ryutaro Tanno, Ira Ktena, et~al.
\newblock Towards generalist biomedical ai.
\newblock {\em Nejm Ai}, 1(3):AIoa2300138, 2024.

\bibitem{chen2023performance}
Yan Chen, Adnan~G Taib, Iain~T Darker, and Jonathan~J James.
\newblock Performance of a breast cancer detection ai algorithm using the
  personal performance in mammographic screening scheme.
\newblock {\em Radiology}, 308(3):e223299, 2023.

\bibitem{wu2019deep}
Nan Wu, Jason Phang, Jungkyu Park, Yiqiu Shen, Zhe Huang, Masha Zorin,
  Stanis{\l}aw Jastrz{\k{e}}bski, Thibault F{\'e}vry, Joe Katsnelson, Eric Kim,
  et~al.
\newblock Deep neural networks improve radiologists’ performance in breast
  cancer screening.
\newblock {\em IEEE transactions on medical imaging}, 39(4):1184--1194, 2019.

\bibitem{chougrad2018deep}
Hiba Chougrad, Hamid Zouaki, and Omar Alheyane.
\newblock Deep convolutional neural networks for breast cancer screening.
\newblock {\em Computer methods and programs in biomedicine}, 157:19--30, 2018.

\bibitem{he2025parameter}
Jinlong He, Pengfei Li, Gang Liu, and Shenjun Zhong.
\newblock Parameter-efficient fine-tuning medical multimodal large language
  models for medical visual grounding.
\newblock In {\em 2025 IEEE 22nd International Symposium on Biomedical Imaging
  (ISBI)}, pages 1--5. IEEE, 2025.

\bibitem{safari2025performancegpt5braintumor}
Mojtaba Safari, Shansong Wang, Mingzhe Hu, Zach Eidex, Qiang Li, and Xiaofeng
  Yang.
\newblock Performance of gpt-5 in brain tumor mri reasoning, 2025.

\bibitem{tajbakhsh2016convolutional}
Nima Tajbakhsh, Jae~Y Shin, Suryakanth~R Gurudu, R~Todd Hurst, Christopher~B
  Kendall, Michael~B Gotway, and Jianming Liang.
\newblock Convolutional neural networks for medical image analysis: Full
  training or fine tuning?
\newblock {\em IEEE transactions on medical imaging}, 35(5):1299--1312, 2016.

\bibitem{wang2025capabilities}
Shansong Wang, Mingzhe Hu, Qiang Li, Mojtaba Safari, and Xiaofeng Yang.
\newblock Capabilities of gpt-5 on multimodal medical reasoning.
\newblock {\em arXiv preprint arXiv:2508.08224}, 2025.

\bibitem{he2020pathvqa}
Xuehai He, Yichen Zhang, Luntian Mou, Eric Xing, and Pengtao Xie.
\newblock Pathvqa: 30000+ questions for medical visual question answering.
\newblock {\em arXiv preprint arXiv:2003.10286}, 2020.

\bibitem{zhang2023pmc}
Xiaoman Zhang, Chaoyi Wu, Ziheng Zhao, Weixiong Lin, Ya~Zhang, Yanfeng Wang,
  and Weidi Xie.
\newblock Pmc-vqa: Visual instruction tuning for medical visual question
  answering.
\newblock {\em arXiv preprint arXiv:2305.10415}, 2023.

\bibitem{jeong2023emory}
Jiwoong~J Jeong, Brianna~L Vey, Ananth Bhimireddy, Thomas Kim, Thiago Santos,
  Ramon Correa, Raman Dutt, Marina Mosunjac, Gabriela Oprea-Ilies, Geoffrey
  Smith, et~al.
\newblock The emory breast imaging dataset (embed): A racially diverse,
  granular dataset of 3.4 million screening and diagnostic mammographic images.
\newblock {\em Radiology: Artificial Intelligence}, 5(1):e220047, 2023.

\bibitem{moreira2012inbreast}
In{\^e}s~C Moreira, Igor Amaral, In{\^e}s Domingues, Ant{\'o}nio Cardoso,
  Maria~Joao Cardoso, and Jaime~S Cardoso.
\newblock Inbreast: toward a full-field digital mammographic database.
\newblock {\em Academic radiology}, 19(2):236--248, 2012.

\bibitem{cui2021chinese}
Chunyan Cui, Li~Li, Hongmin Cai, Zhihao Fan, Ling Zhang, Tingting Dan, Jiao Li,
  and Jinghua Wang.
\newblock The chinese mammography database (cmmd): An online mammography
  database with biopsy confirmed types for machine diagnosis of breast.
\newblock {\em The Cancer Imaging Archive}, 1, 2021.

\bibitem{sawyer2016curated}
Rebecca Sawyer-Lee, Francisco Gimenez, Assaf Hoogi, and Daniel Rubin.
\newblock Curated breast imaging subset of digital database for screening
  mammography (cbis-ddsm).
\newblock {\em (No Title)}, 2016.

\bibitem{openai2025gpt5}
OpenAI.
\newblock Gpt-5.
\newblock \url{https://openai.com/research}, 2025.
\newblock Accessed: 2025-08-14.

\bibitem{chen2024mammo}
Xuxin Chen, Yuheng Li, Mingzhe Hu, Ella Salari, Xiaoqian Chen, Richard~LJ Qiu,
  Bin Zheng, and Xiaofeng Yang.
\newblock Mammo-clip: Leveraging contrastive language-image pre-training (clip)
  for enhanced breast cancer diagnosis with multi-view mammography.
\newblock {\em arXiv preprint arXiv:2404.15946}, 2024.

\bibitem{sun2025multi}
Luhao Sun, Bowen Han, Wenzong Jiang, Weifeng Liu, Baodi Liu, Dapeng Tao,
  Zhiyong Yu, and Chao Li.
\newblock Multi-scale region selection network in deep features for full-field
  mammogram classification.
\newblock {\em Medical Image Analysis}, 100:103399, 2025.

\bibitem{chen2025enhancing}
Han Chen and Anne~L Martel.
\newblock Enhancing breast cancer detection on screening mammogram using
  self-supervised learning and a hybrid deep model of swin transformer and
  convolutional neural networks.
\newblock {\em Journal of Medical Imaging}, 12(S2):S22007--S22007, 2025.

\bibitem{gudhe2024multi}
Naga~Raju Gudhe, Sudah Mazen, Reijo Sund, Veli-Matti Kosma, Hamid Behravan, and
  Arto Mannermaa.
\newblock A multi-view deep evidential learning approach for mammogram density
  classification.
\newblock {\em IEEE Access}, 12:67889--67909, 2024.

\bibitem{lopez2022multi}
Eleonora Lopez, Eleonora Grassucci, Martina Valleriani, and Danilo Comminiello.
\newblock Multi-view hypercomplex learning for breast cancer screening.
\newblock {\em arXiv preprint arXiv:2204.05798}, 2022.

\bibitem{wei2022beyond}
Tao Wei, Angelica~I Aviles-Rivero, Shuo Wang, Yuan Huang, Fiona~J Gilbert,
  Carola-Bibiane Sch{\"o}nlieb, and Chang~Wen Chen.
\newblock Beyond fine-tuning: Classifying high resolution mammograms using
  function-preserving transformations.
\newblock {\em Medical image analysis}, 82:102618, 2022.

\end{thebibliography}

\end{document}